\pdfoutput=1

\documentclass[11pt]{article}

\usepackage[final]{acl}
\usepackage{times}
\usepackage{latexsym}
\usepackage{appendix}
\usepackage{multirow}
\usepackage{graphicx}
\usepackage{amsmath}
\usepackage{nccmath}
\usepackage{algorithm}
\usepackage{algorithmic}

\usepackage[T1]{fontenc}

\usepackage[utf8]{inputenc}

\usepackage{microtype}

\usepackage{inconsolata}

\usepackage{graphicx}

\title{A Data-driven Investigation of Euphemistic Language: \\
Comparing the usage of “slave” and “servant” \\ 
in 19th century US newspapers}

\author{Jaihyun Park \\
  Wee Kim Wee School of \\
  Communication and Information \\
  Nanyang Technological University \\
  \texttt{jay.park@ntu.edu.sg} \\\And
  Ryan Cordell \\
  School of Information Sciences \\
  University of Illinois at \\
  Urbana-Champaign \\
  \texttt{rcordell@illinois.edu} \\}

\begin{document}
\maketitle
\begin{abstract}
\textcolor{red}{Warning: This paper contains examples of offensive language targeting marginalized populations.} 

This study investigates the usage of ``slave'' and ``servant'' in the 19th century US newspapers using computational methods. While both terms were used to refer to enslaved African Americans, they were used in distinct ways. In the Chronicling America corpus, we included possible OCR errors by using FastText embedding and excluded text reprints to consider text reprint culture in the 19th century. Word2vec embedding was used to find semantically close words to ``slave'' and ``servant'' and log-odds ratio was calculated to identify over-represented discourse words in the Southern and Northern newspapers. We found that ``slave'' is associated with socio-economic, legal, and administrative words, however, ``servant'' is linked to religious words in the Northern newspapers while Southern newspapers associated ``servant'' with domestic and familial words. We further found that slave discourse words in Southern newspapers are more prevalent in Northern newspapers while servant discourse words from each side are prevalent in their own region. This study contributes to the understanding of how newspapers created different discourses around enslaved African Americans in the 19th century US.

\end{abstract}

\section{Introduction}

In the United States before the Civil War, free Black Americans and white abolitionists challenged the moral grounds of slavery using a range of means and media, from newspapers that exposed the horrors that enslaved people were subjected to in the American South, to first-hand accounts of former slaves recounting their suffering and their struggle for freedom, to fictional narratives that sought to levy readers’ sympathies toward Black Americans into 
advocacy against slavery. 
Pro-slavery advocates used the same media to undermine abolitionists’ charges and to defend slavery, arguing not simply for its necessity but for its rectitude, even its sanctity. 
Pro-slavery rhetoric often relied on a domestic, sentimental account of slaves’ lives that sidestepped the brutal realities of forced labor by focusing instead on supposed familial bonds between ``house slaves'' and the white women and children who owned them, the religious devotion advocates claimed slavery inculcated in enslaved African Americans, or the reported gratitude of slaves for their condition. 
Such myths underlay everything from newspaper editorials to pro-slavery novels, such as the ``Anti-Tom'' genre that arose to counter the popularity of Harriet Beecher Stowe’s bestseller, \textit{Uncle Tom’s Cabin}. 
In works such as \textit{Uncle Robin in His Cabin in Virginia, And Top Without One in Boston} — the central premise of which is outlined in the title — pro-slavery writers sought to contrast an idyllic depiction of Southern slavery 
— what they euphemistically termed ``our peculiar institution'' — with industrial horrors in the North. 

This research investigates this euphemistic rhetoric through an investigation of two closely linked but rhetorically contrasting terms in Civil War-era US newspapers: slave and servant. Both terms were used to refer to enslaved Black Americans, but they were employed in distinct ways. 
While the word ``slave'' engaged with the slave system directly, marking discourse about legal and political debates around slavery, ``servant'' was more often used euphemistically to identify enslaved people who could be more easily cast into the domestic, sentimental narratives espoused by pro-slavery advocates. 
The latter term can refer to different forms of servitude, including both enslaved workers in the American South and paid domestic help in Northern states, and its use in pro-slavery newspapers to describe enslaved Black Americans deliberately blurred those lines. 
Our goal is not to make claims about the lived experience of Black Americans during this period, but to understand the rhetorical constructions that communicated ideas about enslaved Black Americans to white newspaper readers. 

In newspaper writing, ``slave'' is a more generic word that is used to discuss both enslaved Black Americans, as well as to reference legal and political debates around the institution of slavery itself.
``Servant'' is a more specific term most often used to describe slaves who filled domestic — and less obviously abusive — roles in Southern homes. 
Servants were described as better-dressed and well fed, and typically lived in the attic or basement of their master's house \citep{malcolm1990malcolm}, where they ostensibly enjoyed a better quality of life \citep{gatewood2000aristocrats}.

Any type of text published through newspapers could shape the discourse in the public sphere for editors and readers as newspapers played a role in constituting \textit{imagined communities} in the 19th century \citep{anderson2006imagined}. 
Whether newspapers supported abolition or defended slavery, they nonetheless filtered their understanding of Black Americans through stereotypical filters, while typically excluding Black Americans themselves from their discourse\footnote{For more on Black-owned- and -operated newspapers, see \textit{The Black Newspaper and the Chosen Nation} \citep{fagan2016black}.
In another article \textit{``Chronicling White America,''} \citeauthor{fagan2016chronicling} describes how the collection processes for the Library of Congress's Chronicling America collection have often excluded Black newspapers \citep{fagan2016chronicling}. In part due to this facet of our data, our analyses focus on the stereotypical discourses about Black Americans within white newspapers - to include newspapers that were abolitionist in stance, but run and operated by white editors}. 
In this paper, we present a computational approach to studying the use of these two terms in a corpus of 19th century newspapers and seek to answer two primary research questions:

\begin{itemize}
  \item RQ 1. What are the words that are most similar to ``slave'' and ``servant'' in the corpus of Southern and Northern newspapers?
    \begin{itemize}
      \item RQ 1.1. How do the words that are most similar to ``slave'' and ``servant'' differ among the Northern newspapers and among the Southern newspapers? (Within-newspaper analysis)
      \item RQ 1.2. How do the words that are most similar to ``slave'' and ``servant'' differ between the Southern newspapers and Northern newspapers? (Cross-newspaper analysis)
    \end{itemize}
  \item RQ 2. How prevalent are the discourse words from the Southern and Northern newspapers in the entire corpus? 
\end{itemize}

To support open science and transparent data science, we publish the code used in this study at \url{https://github.com/park-jay/slavery-discourse}.

\section{Related Works}

Scholars of American history, literature, and culture have argued persuasively that while newspapers were not new in the 19th century, they were newly prevalent.
Drawing on data from the Library of Congress's US Newspaper Directory, US newspapers grew ``from a few hundred papers in 1800 to over 12,000 by the end of the century.'' \cite{cordell2020going} 
Beyond the simple scale of this shift, the variety, price, style, and intended audiences for newspapers shifted dramatically during this same period, such that the term ``newspaper'' suddenly encompassed a much wider range of periodicals than existed at the beginning of the century, including merchant papers, penny papers, illustrated family papers, and much more.

This rapidly-growing medium was at the time both highly partisan and strident, as editors debated politicians and each other about political and social issues. 
The voice of newspapers in the 19th century was strongly affiliated with parties and particular political action groups, as modern ideas of journalistic impartiality did not evolve until the early 20th century \citep{pasley_tyranny_2002}. 
Similarly, \citet{baldasty_commercialization_1992} argued newspapers during the antebellum period formed close ties with political parties and factions to gain financial support. Newspapers spoke from the perspective of a ``network author'' that positioned the voice of any individual newspaper within the collective operations of larger political and social discourse \citep{cordell_reprinting_2015}.
Because newspapers have power to frame and manipulate discourse around political and social issues \citep{willaert2022tracking}, researchers have much to learn about how newspaper discourse operated at scale, or how it operated between different regions or time periods.

As \citet{soni2021abolitionist} demonstrate, ``newspapers played a crucial role in spreading information and shaping public opinion about the abolition of slavery and related social justice issues \dots and now serve as a primary source of information about abolition for scholars today.''
This points to a primary challenge facing scholars seeking to use historical newspapers to understand discourses about slavery, as ``those edited by white people (and white men in particular) have been more fully preserved, and therefore, are more accessible to researchers,'' including in most digitized corpora. 

Though some data-driven work on historical newspapers has appeared in recent years, the scale of digital collections argues for more research that will supplement the findings and analysis from qualitative scholars \citep{gabrial2004melancholy,narayan_slavery_2020}. 
Scholars in digital humanities have used digitized texts to study culture \cite{griebel2024locating}, applying computational models built on diverse datasets \cite{park2022raison} and to explore historical change over time. 
Especially, newspapers as a source of data attracted computational humanities researchers and they leveraged computational methods \citep{park2023quantitative} such as topic modeling \cite{hengchen_data-driven_2021,klein_exploratory_2015} and word embedding models \cite{soni2021abolitionist} to learn about the past. 


More relevant to our study, \citet{gabrial2004melancholy} claimed that there is discourse in newspapers around the idea of ``a good negro'' by showing cases where newspapers reported that they could stop race riots thanks to ``loyal'' Black Americans (e.g., ``a servant prompted by attachment to his master revealed the conspiracy'', ``some faithful Blacks had informed the Charleston City Council'' p.310). 
The work of \citet{gabrial2004melancholy} illustrates that newspapers framed accounts of Black Americans to maintain white supremacy.

In this study, we seek not to read between the lines of predominantly white-edited newspapers to identify secretly liberatory language, but instead to read directly, using computational methods, the way that white editors on both sides of the political spectrum, from the abolitionist (e.g.,textit{Anti-Slavery Bugle} (New Lisbon, Ohio)) to the pro-slavery (e.g., \textit{Daily Dispatch} (Richmond, Virginia)), deployed contrasting stereotypes of Black Americans for their own rhetorical purposes. 
We explore how ``slaves'' and ``servants'' — words which are shorthand for broader discourses — were discussed in newspapers between the introduction of the Fugitive Slave Act of 1850 and the end of the Civil War, and then compare that discourse in abolitionist newspapers, largely located in the North, to discourse in pro-slavery Southern papers. 
By approaching racial bias embedded in language use within newspapers, rather than trying to identify justifications for stereotypical rhetoric, we seek to use computing ``as a diagnostic, helping us to understand and measure social problems with precision and clarity,'' as well ``as synecdoche'' that ``makes long-standing social problems newly salient in the public eye''\citep{abebe2020roles}.

\section{Data} \label{data}
\subsection{Data Collection} \label{data collection}

Through digitization, many archival materials have been made available to the public \cite{dobreski2020remodeling}.
Among many available public datasets for 19th century newspapers, we used Chronicling America as a source for data collection. 
We did not use any other sources in order to keep consistency in OCR errors (Some digitized newspapers have article breaks, meaning document is article-level but Chronicling America does not have article breaks, having digitization at page-level.).
Of data available in Chronicling America, if it is digitized through a high-resolution, lossless digital image of a microfilm copy, then the quality of data is relatively reliable \citep{lorang_electronic_2012}. 
If we complement the dataset with perfectly transcribed digital text or use computer vision to create our own dataset, then OCR errors are not controlled in the word embedding model. 
Biases from the computer vision algorithm could have impacted the quality of the word embedding model that worked differently from OCR error from Chronicling America. 

The cause of the Civil War is in part because of the dispute and disagreement of maintaining slavery institution. 
The Fugitive Slave Act of 1850 empowered the Federal government to intervene in the legal issues that may arise when the slave escaped to the Free State. 
On the surface, the Fugitive Slave Act represents a \textit{de jure} improvement in slave owners’ property rights \citep{lennon_slave_2016}. 
However, the Fugitive Slave Act complicated the problem of slavery by creating a conflict of jurisdictions between the States  and the Federal government \citep{baumgartner_enforcing_2022}. 
The election of Abraham Lincoln catalyzed the secession of the Southern States even though the Fugitive Slave Act compromised the Southern interest of maintaining \textit{status quo} of slavery and the Northern interest of the abolitionist movement. 
In order to capture the period of the Fugitive Slave Act and the period of the Civil War, we collected data from January 1st, 1850 to December 31st, 1865.
There are 3,803 digitized newspapers in Chronicling America\footnote{\url{https://chroniclingamerica.loc.gov/} searched on September 24th, 2022.}.
Neither all of them are about abolitionist movement nor written in English, we chose 7 newspapers for our study using details in the newspaper biographies provided by Chronicling America.

We wanted the newspapers to have (1) enough data to cover the date we want to study, (2) regional representativeness (both Southern and Northern newspapers), and (3) identified as abolitionist or anti-abolitionist newspapers, following the stance of editors.  
We included the two well-known abolitionist newspapers (\textit{Anti-slavery Bugle} and \textit{National Era}) in addition to one anti-slavery newspaper from New England (\textit{Green Mountain Freeman}) and four deep South newspapers (\textit{Abbeville Banner}, \textit{Daily Dispatch}, \textit{Edgefield Advertiser}, and \textit{Nashville Union and American}). 
Arguably, Southern papers we included in the study were not designated for anti-abolition movement whereas \textit{Anti-Slavery Bugle} and \textit{National Era} advocated the abolitionist movement. 
However, as Southern states saw abolition could lead to a financial crisis, editors were, in most cases, anti-abolitionists. 
The detail of information of our newspapers is provided in Table \ref{tab:data} in the appendix.

\subsection{Data preparation} \label{data preparation}
19th century newspapers have multiple columns \citep{smith_computational_2015}.
This made Chronicling America scan the entire page of the newspaper rather than splitting the text based on sections or articles. 
Unfortunatley, this characteristic poses difficulties in processing the text. 
Our study focuses on specific words (i.e., ``slave'' and ``servant''),
so building embeddings from the entire page would introduce unnecessary information into the model.
In addition, the poor quality of OCRed text provides a wrong position of where the sentence ends. Therefore, as a way of reduce the problems stated above, we read the OCR text by line and took where the word of interest appeared.

As a way of addressing this problem, we identified passages where our keywords (``slave'' or ``servant,'' respectively) appeared, and then defined a window of two lines before and after where the keyword appeared. 
While this method does not perfectly align with article breaks, it does capture more of the context around our keywords without including the full text of the page, which might include significant amounts of text that is entirely irrelevant to the topics we are studying. 
These snippets are the primary data studied using the methods outlined below. 
Once we identified the index of the line where the word of interest appeared, we took the subset of the data with one line before and after to make a snippet. 
This is to ensure we have enough contextual words around the word of interest at best. 

\section{Methodology} \label{methodology}

\subsection{FastText cosine similarity} \label{fasttext cosine similarity}
It is well-known that OCR produces wrong predictions when the scanned page is worn or damaged. OCR errors can bias the results \citep{chiron_impact_2017} and scholars are often uncertain whether the result is publishable \citep{traub_impact_2015}.
The error introduced by OCR can impact the overall quality of research as well as the performance of the NLP model \citep{jiang_impact_2021}. 
The more the OCR produced errors included in the text, the more the word tokens that the NLP model recognizes. 

Especially when we use the Bag-of-Words (BoW) approach to conduct research, it is recommended to handle OCR errors due to the possibility of misrepresentation of actual word counts and OCRed word counts. 
For example, even though humans can interpret ``slove'' as ``slave'' considering that ``slove'' might be OCR error, the BoW approach will take ``slove'' as a unique word and index it. 
This will eventually expand the list of words and distort the real distribution of words in a given corpus.

In order to reduce this negative influence of OCR errors, we used FastText embedding \citep{bojanowski_enriching_2017} to identify possible candidates for OCR errors. 
Since FastText takes character-level n-grams instead of word-level embedding to build context-aware embeddings, it is tested effective that FastText can generate possible candidates for OCR errors \citep{hajiali_generating_2022}. 
Therefore, we listed the most similar words based on Cosine distance metrics to the words of interest using FastText embedding. 
Finally, human annotators coded whether the words listed as similar words could be considered as OCR errors. 

\subsection{Human annotation decision process} \label{human annotation decision process}

\begin{algorithm}[h]
  \raggedright
  \caption{Human annotation process}\label{alg:human}
  \begin{algorithmic}[1]
  \STATE $dic$ $\leftarrow$ words in dictionary
  \STATE $sem$ $\leftarrow$ words semantically relevant
  \STATE $par$ $\leftarrow$ words with missing characters
  \STATE $fun$ $\leftarrow$ function words
  \STATE $gen$ $\leftarrow$ gender words
  \STATE $data$ $\leftarrow$ FastText Cosine similarity results
  \STATE $list$ $\leftarrow$ \{\}
  \FORALL{element $\in$ $data$}
    \IF{element $\in$ $dic$ \AND element $\notin$ $sem$}
      \STATE \textbf{continue}
    \ELSIF{element $\in$ $par$}
      \STATE \textbf{continue}
    \ELSIF{element $\notin$ $par$ \AND element $\in$ $fun$}
      \STATE $list$ $\leftarrow$ element
    \ELSIF{element $\notin$ $par$ \AND element $\in$ $gen$}
      \STATE $list$ $\leftarrow$ element
    \ENDIF
  \ENDFOR
  \STATE \textbf{return} $list$
  \end{algorithmic}
\end{algorithm}

Two human annotators (One is from information science background and the other is from literary studies background) read samples of OCR errors and coded without any discussion. 
We had binary categories, which were ```include'' if it can be considered as OCR error or ``exclude'' if it is not considered to be OCR error. 
Our first round of intercoder reliability measure \citep{cohen_coefficient_1960}
for ``slave'' and ``servant'' were substantial ($k$=0.73) and fair ($k$=0.39) agreement respectively \citep{viera_understanding_nodate}. 
Two annotators discussed why disagreement arose. FastText returned ``slavery'' as a culture of practicing enslavement of Black and it was a mix of ``slavery'' OCR errors and ``slave'' OCR errors. 

Therefore, we set more specific rules to decide what to include and what to exclude and the logistics are provided in Algorithm \ref{alg:human}. Two annotators conceptualized $dic$ as the words appeared in dictionaries, $sem$ as the words semantically relevant words (e.g., ``slavery'') and $par$ as the words with missing characters (e.g., ``slav'', ``serva''), $fun$ as the function words which does not have semantical information as well as the characters (e.g., ``of'', ``for'', ``t''), and $gen$ as gender words (e.g., ``man'', ``woman''). 
The human annotating process started with reading through the FastText Cosine similarity results and if the word can be found in the dictionary and not relevant to the word of interest (e.g., ``hold'', ``buy''), we excluded the word. 
In the next step, two annotators checked whether the word has partial characters of the word of interest (e.g., ``sla''), if this was the case, the word was excluded because the actual word of it could have been  other words in dictionaries (e.g., ``slate'', ``slam''). 
In order to reduce false positives as many as possible and to make the logic of choosing OCR error candidates as less greedy as possible, the words with partial characters were excluded. 

If the word from FastText entailed the extra characters on the tail (e.g., ``slaveto'', ``slavewith'') and they were the function words which does not convey information in terms of semantics, the annotators regarded it as OCR errors with tokenization and included in the OCR error candidates. 
Last but not least, annotators checked whether the words attached the gender-related words on the tail (e.g., ``slaveman'', ``servantwoman'') and if this was the case, we included as one of the possible OCR error candidates. 

The average Levenshtein distance \citep{levenshtein_binary_1966} of the OCR error candidates for ``slave'' is 75.89 while ``servant'' is 80.48. 
The standard deviation for ``slave'' is 9.92 when ``servant'' is 8.81. 
With these OCR candidates, we added the snippet of OCR candidates and the final size of dataset for this study is presented in Table \ref{tab:data}. 
Overall, the size of snippets after including OCR error candidates increased 5,765 for servant data (1.15\%) and 14,241 for slave data (1.07\%).    

\subsection{Text reprints deduplication} \label{text reprints deduplication}
19th century American newspapers reprinted texts from a wide range of genres: news reports, recipes, trivia, lists, vignettes, and religious reflections \citep{cordell__2017}. 
Text reprints could also include boilerplate that appeared across many issues of the same paper, such as advertisements. 
A business might buy advertisement space for multiple weeks, months, or even years, and those ads would be left in standing type from issue to issue. 

In a study such as this one, focused on textual reuse, an ad that includes a keyword of interest but which appears day after day can disproportionately influence the statistical relationship between words in the corpus, leading our model to overestimate the importance of words within the ad relative to the words in texts that changed each day. In other words, if one particular phrase repeatedly appears, then the embedding model will overfit the phrase because of the distorted distribution of the text. However, it is hard to detect reprints based on keyword searches because of OCR errors. 

Here we adopt the text-reuse detection methods, as described in \citet{smith_detecting_2014}, which use n-gram document representations to detect text reprints within errorful OCR-derived text. 
We processed our corpus with a 5-gram chunking using NLTK whitespace tokenizer and further made a judgment that the text has been reprinted when there were more than three matches of 5-grams across the snippets. 
For deduplication, we kept only the first snippet among multiple reprints. 

For instance, the advertisement about selling a servant with the detailed condition appeared in \textit{Daily Dispatch} on January 9th, 1856; 
(e.g., ``child'', ``4'', ``12'', ``year'', ``age'', ``ser'', ``vant'', ``half'', ``price'', ``servant'', ``travel'', ``bv'', ``must'', ``furnish'', ``two'', ``pass'', ``one'', ``may'') 
was detected to have reprints in May 20th, 1856 
(e.g., ``child'', ``4'', ``12'', ``year'', ``age'', ``ser'', ``vant'', ``half'', ``price'', ``servant'', ``travel'', ``must'', ``famish'', ``two'', ``pass'', ``one'', ``may''). 
In this example, this pair is not identical because of inconsistent OCR like ``furnish'' and ``famish'', however, the 5-gram matching examination substantiated that this pair denotes a reprinted text. 
Due to the effectiveness of the method by \citet{smith_detecting_2014}, we used the method of n-gram document representations to prepare the final data for a Word2vec model. 
After the deduplication process, the size of the snippets decreased by 13,533 in servant data (0.68\%) and 23,089 data in slave data (0.88\%). 
The final size used for the analysis is provided in Table \ref{tab:data}.

\subsection{Word2vec embedding} \label{word2vec embedding}
Once we prepared snippets of the text where ``slave'' and ``servant'' including possible OCR error candidates and excluding text reprints, we trained Word2vec model \citep{mikolov_distributed_2013} to leverage CBOW (Continuous Bags of Words) and Skip-gram model.
As an initial step for preparing the training snippets, we used the standard stopwords list from NLTK\footnote{\url{https://www.nltk.org/}}.
Once we got rid of the words from stopwords list, we then made the word lower case and then we lemmatized the words by relying on \textit{en\_core\_web\_sm} model from SpaCy\footnote{\url{https://spacy.io/models}}.
To make sure that the embedding model does not overfit miscellaneous OCR errors, we had the embedding model train only the words that appeared more than 10 times in the entire snippets.

\subsection{Statistically over-represented discourse words} \label{over-represented discourse words}
We operationalized the discourse words as the words that are the words close to ``slave'' and ``servant'' from Word2vec embedding (sec \ref{word2vec embedding}).
In order to answer RQ2, where we address how the words close to ``slave'' and ``servant'' prevalent in the entire corpora of newspapers, we calculated log-odds ratio with informative Dirichlet as defined in equation \ref{eq: log-odds ratio} \citep{monroe2008fightin}.

\begin{equation}
\begin{aligned}
  \delta_{w}^{(i-j)} = \log \frac{y_{w}^{i}+a_{w}}{n^{i}+a_{0}-y_{w}^{i}-a_{w}} \\
  - \log \frac{y_{w}^{j}+a_{w}}{n^{j}+a_{0}-y_{w}^{j}-a_{w}}
\end{aligned}
\label{eq: log-odds ratio}
\end{equation}

The log-odds ratio with informative Dirichlet of each word $w$ between two corpora $i$ and $j$ (in our study, newspapers from the North and the South) given the prior frequencies are obtained from the entire corpus $a$. When $n^{i}$ is the total number of words in corpus $i$, $y_{w}^{i}$ is the number of times word $w$ appears in corpus $i$, $a_{0}$ is the size of the corpus $a$, and
$a_{w}$ is the frequency of word $w$ in corpus $a$ \citep{kwak2020systematic}.
With the log-odds ratio, we can identify the words that are over-represented 
in the corpora \citep{park2023quantitative, park2024you}.

\section{Findings} \label{Findings}


\subsection{RQ1: What are the words that are similar words?} \label{findings:rq1}
In general, we found that discourse around slave (Table \ref{tab:my-result-slave}) is centered around socio-economic, legal, and administrative words, regardless of the source newspaper's stance toward slavery.
By contrast, discourse around servant (Table \ref{tab:my-result-servant}) from pro-slavery stance newspapers is more related to domestic work whereas discourse around servant from anti-slavery stance newspapers is mostly comprised by religious words.

Socio-economic, legal, and administrative words are prevalent in slave discourse compared to servant discourse.
For instance, ``congress'' (from \textit{Edgefield Advertiser} and \textit{Green Mountain Freeman}), ``constitution'' (from \textit{Edgefield Advertiser} and \textit{Green Mountain Freeman}), ``legislate'' (from \textit{Edgefield Advertise} and \textit{Green Mountain Freeman}), ``nation'' (from \textit{Abbeville Banner}), and ``commonwealth'' (from \textit{Daily Dispatch}) can be considered words with legal and administrative implications.

Words such as ``attempt,'' ``death,'' ``punishment,'' ``violation,'' and ``crime,'' might be pertinent to the frame that pro-slavery newspaper had tried to scheme.
Similarly, we can also observe ``fugitive'' from \textit{Edgefield Advertiser}.
However, \textit{Daily Dispatch} contains more words implying negative or violent actions from slaves than any other newspapers.

We also able to observe that economic words like ``profit'' (From \textit{Abbeville Banner}) and ``labor'' (From \textit{Abbeville Banner} and \textit{Nashville Union and American}).
In addition, the words around how to rebuild the nation after the Civil War is also captured in \textit{Anti-slavery Bugle} (``restoration'')

Contrary to slave discourse, which was largely unanimous across pro-slavery and anti-slavery newspapers, servant discourse was starkly divided by the newspapers' stance toward slavery.

While pro-slavery newspapers from the South showed more words around domesticity, anti-slavery newspapers showed more religiously-inflected words.
We can find words like ``table,'' ``ice,'' ``furniture,'' ``garden,'' ``ladle,'' and ``dress'' closely aligned with "servant" in the \textit{Daily Dispatch}, ``seat,'' ``cook,'' and ``house,'' in the \textit{Edgefield Advertiser}, ``linen,'' ``bed,'' ``shirt,'' ``dress,'' ``cook,'' ``flannel,'' ``apron,''  ``cottonade,'' and ``blanket'' in the \textit{Nashville Union and American}, all pro-slavery newspapers. 

In addition, we observed that the words associated with a good demeanor that conforms to white supremacist societal hierarchies appeared regardless of the papers' stance toward slavery.
For instance, ``respectfully'' (from \textit{Edgefield Advertiser}, \textit{Green Mountain Freeman}, and \textit{National Era}) ``obedient'' (from \textit{Edgefield Advertiser}, \textit{Green Mountain Freeman}, and \textit{National Era}), and ``humble'' (from \textit{National Era}).

However, religiously laden words are unique to anti-slavery newspapers. The words like ``bible,'' ``jesus,'' (from \textit{Anti-slavery Bugle}) ``god'' (from \textit{Green Mountain Freeman} and \textit{National Era}), ``christ,'' (\textit{Anti-slavery Bugle} and \textit{National Era}), ``faithful'' (\textit{National Era}) can be religious words.

In summary, we find that discourse around the word ``slave'' is more focused on macroscopic concepts including socio-economic, legal, and administrative words compared to servant discourse (RQ1-1). 
Although we cannot find a stark contrast between slave discourse in pro-slavery newspapers and anti-slavery newspapers, we observe that there is a difference between pro-slavery newspapers (religious accounts) and anti-slavery newspapers (domestic work words) in servant discourse (RQ1-2).

\subsection{RQ2: How prevalent are the discourse words?} \label{findings:rq2}
With the findings in section \ref{findings:rq1}, we explored how prevalent the discourse words are in the corpus. 
If the datapoint is above 0 in the Y-axis, it means that the word is over-represented in the Northern newspapers and if the datapoint is below 0 in the Y-axis, it means that the word is over-represented in the Southern newspapers. 

\begin{figure*}[htb]
  \centering
  \includegraphics[width=1\textwidth]{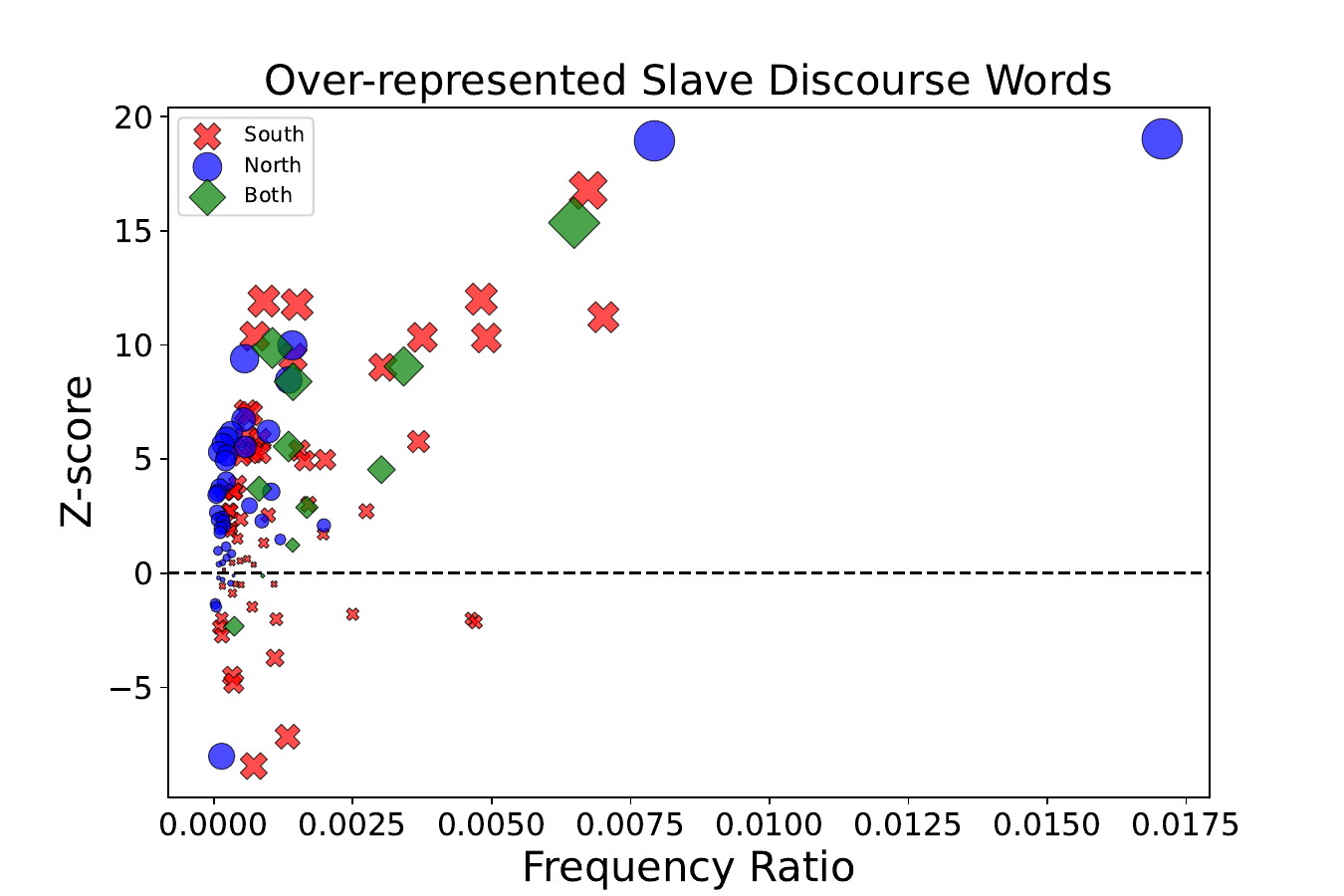}
    \caption{The datapoints represent the slave discourse words. The slave discourse words from the South is represented with red cross, the slave discourse words from the North is represented with blue circle, and the slave discourse words that appeared in both Southern and Northern newspapers are in square diamond with green color. X-axis shows the frequency of the words in the entire corpus
    and Y-axis shows the Z-score of the words in the corpus.
    }~\label{fig:slave-discourse}
    \vspace{-4mm}
  \end{figure*}

In figure \ref{fig:slave-discourse}, we can observe that 45 slave discourse words from the South are over-represented in the Northern newspapers (0.7142\% of the South slave discourse words) while only 7 slave discourse words from the North are over-represented in the Southern newspapers (0.1627\% of the North slave discourse words). 

This indicates that the slave discourse words are more used in the Northern newspapers than the Southern newspapers. The words like ``law'' (Z=16.7714), ``trade'' (Z=12.0071), ``nation'' (Z=11.9236) are the top over-represented slave Southern discourse words in the Northern newspapers. 
On the other hand, ``pro'' (Z=$-$0.84507), ``owner'' (Z=$-$7.1651), ``death'' (Z=$-$4.8179) are the top over-represented slave Southern discourse words in the Southern newspapers. 

This finding is in part explainable by the findings from the RQ 1 where we found that slave discourse is more focused on macroscopic concepts including socio-economic, legal, and administrative words. 
Since these words were also used in describing the slave in Northern newspapers, we can observe that the words from the South are also frequently used in the Northern newspapers. 

\begin{figure*}[htb]
    \centering
    \includegraphics[width=1\textwidth]{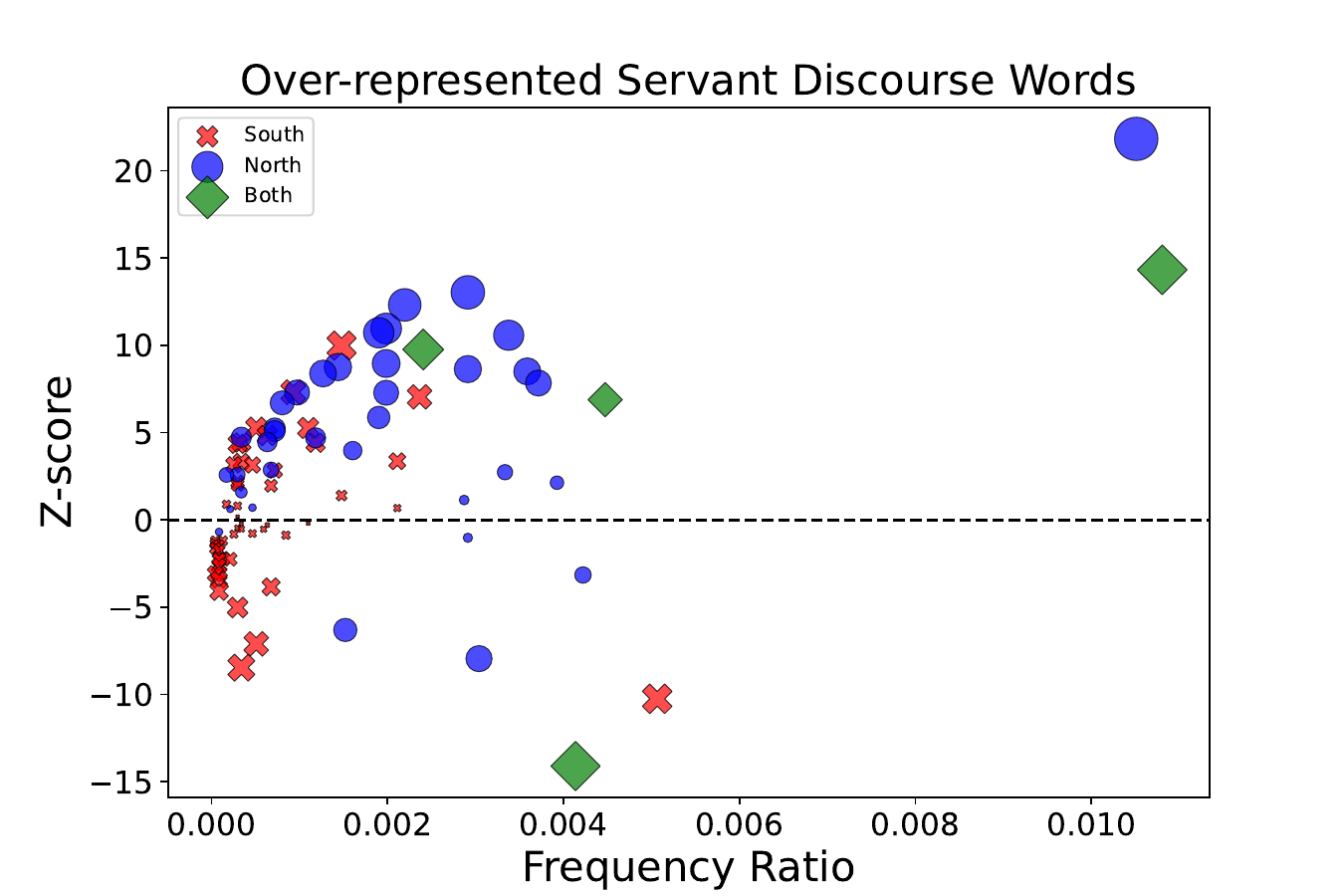}
    \caption{The datapoints represent the servant discourse words. The servant discourse words from the South is represented with red cross, the slervant discourse words from the North is represented with blue circle, and the servant discourse words that appeared in both Southern and Northern newspapers are in square diamond with green color. X-axis shows the frequency of the words in the entire corpus
    and Y-axis shows the Z-score of the words in the corpus.
    }~\label{fig:servant-discourse}
    \vspace{-4mm}
  \end{figure*}

However, as we found in RQ 1, the servant discourse words showed contrast between the Northern and Southern newspapers. 
This is also reflected in the word usage in the Northern and Southern newspapers. 
From figure \ref{fig:servant-discourse}, we can observe that the majority of the servant discourse words from the South and the servant discourse words from the North are over-represented in the newspapers they are from. 

29 servant discourse words from the South are frequently used in the Southern newspapers (0.5576\% of the Southern servant discourse words). 
Similarly, 31 servant discourse words from the North are freqeucntly used in the Northern newspapers (0.8611\% of the Northern servant discourse words).
The words related to religion such as ``faithful'' (Z=8.5067), ``christ'' (Z=5.2329), ``church'' (Z=0.6994) are more frequently used in the Northern newspapers than the Southern newspapers. The words characterized by domestic work such as ``apron'' (Z=$-$1.6263), ``shirt'' (Z=$-$2.2580), ``linen'' (Z=$-$3.4279), and ``cook'' (Z=$-$7.1075) are more frequently used in the Southern newspapers than the Northern newspapers.

\section{Discussion} \label{discussions}
For Southern editors and readers, slaves were property which could be taken away through political action and this concern was well-reflected in the \textit{Abbeville Banner}. 
Indeed, the Southern economic system cannot be explained without institutionalized slavery \citep{meyer_economics_2017}. 
For them, slave labor undergirded and sustained the Southern economy. 
Because the role of slave is deeply connected to economy and society, discourse around slave in 19th century newspapers, at least during the period of our data, is mostly centered around law and government. 
Unfortunately, slave discourse from newspapers does not show how subjugated life of slave actually was. Even for anti-slavery stance newspapers does not frame slavery matter with empathy-provoking words to speak more audiences. 

We also found that the slave discourse words from the South are over-represented in Northern newspapers.
We hypothesize that this is because the slave discourse words are based on the political, legal, and economic situations of the United States. 
Since the discourse around slave-related words from the Northern newspapers discusses slaves in a similar manner, the words from the South are also frequently used in Northern newspapers.

Contrary to iron-hearted accounts for slaves, servant discourse contains words for family, and every-day life of servants. 
In addition to domestic work of the duty of servants, words related to family can be emotional. 
Taking this together, the sentence combined with ``respectfully'', ``obedient'', servants ``cook''ing for a hot soup is sufficient enough to imagine warm hospitality and thus evoke nostalgic imagination of South \citep{mcpherson_reconstructing_2003}. 

By demonstrating that discourse around ``servant'' in Southern newspapers euphemized and idealized the depiction of slavery, our findings can supplement the work of \citet{glazer_carry_1996} which studied nostalgia for an idyllic antebellum South in 19th century popular culture.
Even though \textit{Gone with the Wind}, published in 1936, is attributed to the claim of re-construction of nostalgic South, we observed the emergence of early prototype of creating nostalgic South by associating ``servants'' with sentimental and patronizing words. 

We find that Southern newspapers were far more likely to use words that created a sentimental or nostalgic image of the South and the slave system.
Servant discourse words from the South are not frequently used in the Northern newspapers. This uniqueness helps explain how the word ``servant'' is used euphemistically to describe domestic slaves in the South, downplaying their forced servitude by using more neutral, domestic words. 
Though on the surface, ``servant'' might seem like a more benign and positive word than ``slave'', the patterns of word usage in the newspapers suggest how Southern newspapers language worked rhetorically to stereotype Black Americans and sanitize the brutal system of oppression and subjugation.

Abolitionist newspapers relied on evangelical rhetoric to discuss servants compared to slaves. Though biblical justifications were often used to defend slavery (i.e., Genesis 9:18-27; Ephesians 6:5-7), the abolitionist movement also drew hevily on religious conviction and language in articulating the case for emancipation \citep{rae_great_2018}.

It resonates the historical context that the emphasis on Bible has led North to include the feminist and temperance movements by marring the integrity of Biblical authority while helped South to revive religious spirit \citep{lloyd_slavery_1939}. 

In Imagined Communities, \citet{anderson2006imagined} said ``... the very conception of the newspaper implies the refraction of even `world events' into a specific imagined world of vernacular readers.'' In other words, newspapers are not only a reflection of the society but also a tool to shape the society. The interplay between the newspapers creating the nostalgic image of the South and the society that consumed the newspapers led to a reinforcement of the sentimental and idealized portrayal of slavery. This cyclical relationship between media and society calls for a more research on how the past was 
shaped by the media and how it influenced the public perception of slavery.

This study adds to the scholarship on digital humanities by providing a computational approach to understanding how 19th century newspapers framed the discourse around "slave" and "servant." By leveraging word embeddings and statistical analysis, we were able to uncover the nuanced differences in how these terms were used in pro-slavery and anti-slavery newspapers. Our findings highlight the role of language in shaping public perception and the importance of critically examining historical texts to understand the socio-political context of the time.

\section{Acknowledgements} \label{acknowledgements}
Authors would like to thank Matthew Kollmer, the PhD student in the School of Information Sciences at the University of Illinois at Urbana-Champaign, for his insightful comments on the manuscript and participating in the annotation process. 
This research was in part supported by Eugene Garfield Doctoral Dissertation Fellowship 2024 from the Beta Phi Mu International Library and Information Studies Honor Society.

\bibliography{custom}

\begin{thebibliography}{43}
\providecommand{\natexlab}[1]{#1}

\bibitem[{Abebe et~al.(2020)Abebe, Barocas, Kleinberg, Levy, Raghavan, and Robinson}]{abebe2020roles}
Rediet Abebe, Solon Barocas, Jon Kleinberg, Karen Levy, Manish Raghavan, and David~G Robinson. 2020.
\newblock Roles for computing in social change.
\newblock In \emph{Proceedings of the 2020 conference on fairness, accountability, and transparency}, pages 252--260.

\bibitem[{Anderson(2006)}]{anderson2006imagined}
Benedict Anderson. 2006.
\newblock \emph{Imagined communities: Reflections on the origin and spread of nationalism}.
\newblock Verso books.

\bibitem[{Baldasty(1992)}]{baldasty_commercialization_1992}
Gerald~J. Baldasty. 1992.
\newblock \emph{The {Commercialization} of {News} in the {Nineteenth} {Century}}.
\newblock Univ of Wisconsin Press.
\newblock Google-Books-ID: eCG98jIAG\_MC.

\bibitem[{Baumgartner(2022)}]{baumgartner_enforcing_2022}
Alice~L. Baumgartner. 2022.
\newblock Enforcing the {Fugitive} {Slave} {Acts} in the {South}: {Federalism}, {Irony}, and the {Conflict} of {Jurisdictions}, 1787–1861.
\newblock \emph{Journal of Southern History}, 88(3):475--500.
\newblock Publisher: The Southern Historical Association.

\bibitem[{Bojanowski et~al.(2017)Bojanowski, Grave, Joulin, and Mikolov}]{bojanowski_enriching_2017}
Piotr Bojanowski, Edouard Grave, Armand Joulin, and Tomas Mikolov. 2017.
\newblock Enriching word vectors with subword information.
\newblock \emph{Transactions of the association for computational linguistics}, 5:135--146.
\newblock Publisher: MIT Press.

\bibitem[{Chiron et~al.(2017)Chiron, Doucet, Coustaty, Visani, and Moreux}]{chiron_impact_2017}
Guillaume Chiron, Antoine Doucet, Mickaël Coustaty, Muriel Visani, and Jean-Philippe Moreux. 2017.
\newblock Impact of {OCR} errors on the use of digital libraries: towards a better access to information.
\newblock In \emph{2017 {ACM}/{IEEE} {Joint} {Conference} on {Digital} {Libraries} ({JCDL})}, pages 1--4. IEEE.

\bibitem[{Cohen(1960)}]{cohen_coefficient_1960}
Jacob Cohen. 1960.
\newblock A coefficient of agreement for nominal scales.
\newblock \emph{Educational and psychological measurement}, 20(1):37--46.
\newblock Publisher: Sage Publications Sage CA: Thousand Oaks, CA.

\bibitem[{Cordell(2015)}]{cordell_reprinting_2015}
Ryan Cordell. 2015.
\newblock Reprinting, circulation, and the network author in antebellum newspapers.
\newblock \emph{American Literary History}, 27(3):417--445.
\newblock Publisher: Oxford University Press.

\bibitem[{Cordell and Mullen(2017)}]{cordell__2017}
Ryan Cordell and Abby Mullen. 2017.
\newblock " {Fugitive} {Verses}”: {The} {Circulation} of {Poems} in {Nineteenth}-{Century} {American} {Newspapers}.
\newblock \emph{American Periodicals}, 27(1):29--52.
\newblock Publisher: JSTOR.

\bibitem[{Cordell et~al.(2020)Cordell, Smith, Mullen, Fitzgerald, and Kinias}]{cordell2020going}
Ryan Cordell, David~A Smith, Abby Mullen, Jonathan~D Fitzgerald, and T~Kinias. 2020.
\newblock \emph{Going the Rounds: Virality in Nineteenth-Century American Newspapers}.
\newblock University of Minnesota Press, Forthcoming.

\bibitem[{Dobreski et~al.(2020)Dobreski, Park, Leathers, and Qin}]{dobreski2020remodeling}
Brian Dobreski, Jaihyun Park, Alicia Leathers, and Jian Qin. 2020.
\newblock Remodeling archival metadata descriptions for linked archives.
\newblock In \emph{Proceedings of the international conference on dublin core and metadata applications}. Dublin Core Metadata Initiative.

\bibitem[{Fagan(2016{\natexlab{a}})}]{fagan2016black}
Benjamin Fagan. 2016{\natexlab{a}}.
\newblock \emph{The Black Newspaper and the Chosen Nation}.
\newblock University of Georgia Press.

\bibitem[{Fagan(2016{\natexlab{b}})}]{fagan2016chronicling}
Benjamin Fagan. 2016{\natexlab{b}}.
\newblock Chronicling white america.
\newblock \emph{American Periodicals: A Journal of History \& Criticism}, 26(1):10--13.

\bibitem[{Gabrial(2004)}]{gabrial2004melancholy}
Brian~Ray Gabrial. 2004.
\newblock \emph{“The melancholy effect of popular excitement”: Discourse about slavery and the social construction of the slave rebel and conspirator in newspapers}.
\newblock Ph.D. thesis, University of Minnesota.

\bibitem[{Gatewood(2000)}]{gatewood2000aristocrats}
Willard~B Gatewood. 2000.
\newblock \emph{Aristocrats of color: The black elite, 1880--1920}.
\newblock University of Arkansas Press.

\bibitem[{Glazer and Key(1996)}]{glazer_carry_1996}
Lee Glazer and Susan Key. 1996.
\newblock Carry me back: {Nostalgia} for the old {South} in nineteenth-century popular culture.
\newblock \emph{Journal of American Studies}, 30(1):1--24.
\newblock Publisher: Cambridge University Press.

\bibitem[{Griebel et~al.(2024)Griebel, Cohen, Li, Park, Liu, Perkins, and Underwood}]{griebel2024locating}
Sarah Griebel, Becca Cohen, Lucian Li, Jaihyun Park, Jiayu Liu, Jana Perkins, and Ted Underwood. 2024.
\newblock Locating the leading edge of cultural change.
\newblock In \emph{CHR 2024: Computational Humanities Research Conference}, pages 232--245.

\bibitem[{Hajiali et~al.(2022)Hajiali, Fonseca~Cacho, and Taghva}]{hajiali_generating_2022}
Mahdi Hajiali, Jorge~Ramón Fonseca~Cacho, and Kazem Taghva. 2022.
\newblock Generating {Correction} {Candidates} for {OCR} {Errors} using {BERT} {Language} {Model} and {FastText} {SubWord} {Embeddings}.
\newblock In \emph{Intelligent {Computing}: {Proceedings} of the 2021 {Computing} {Conference}, {Volume} 1}, pages 1045--1053. Springer.

\bibitem[{Hengchen et~al.(2021)Hengchen, Ros, Marjanen, and Tolonen}]{hengchen_data-driven_2021}
Simon Hengchen, Ruben Ros, Jani Marjanen, and Mikko Tolonen. 2021.
\newblock A data-driven approach to studying changing vocabularies in historical newspaper collections.
\newblock \emph{Digital scholarship in the humanities}, 36(Supplement\_2):ii109--ii126.
\newblock Publisher: Oxford University Press.

\bibitem[{Jiang et~al.(2021)Jiang, Hu, Worthey, Dubnicek, Underwood, and Downie}]{jiang_impact_2021}
Ming Jiang, Yuerong Hu, Glen Worthey, Ryan~C. Dubnicek, Ted Underwood, and J.~Stephen Downie. 2021.
\newblock Impact of {OCR} {Quality} on {BERT} {Embeddings} in the {Domain} {Classification} of {Book} {Excerpts}.
\newblock In \emph{{CHR}}, pages 266--279.

\bibitem[{Klein et~al.(2015)Klein, Eisenstein, and Sun}]{klein_exploratory_2015}
Lauren~F. Klein, Jacob Eisenstein, and Iris Sun. 2015.
\newblock Exploratory thematic analysis for digitized archival collections.
\newblock \emph{Digital scholarship in the humanities}, 30(suppl\_1):i130--i141.
\newblock Publisher: Oxford University Press.

\bibitem[{Kwak et~al.(2020)Kwak, An, and Ahn}]{kwak2020systematic}
Haewoon Kwak, Jisun An, and Yong-Yeol Ahn. 2020.
\newblock A systematic media frame analysis of 1.5 million new york times articles from 2000 to 2017.
\newblock In \emph{Proceedings of the 12th ACM Conference on Web Science}, pages 305--314.

\bibitem[{Lennon(2016)}]{lennon_slave_2016}
Conor Lennon. 2016.
\newblock Slave escape, prices, and the fugitive slave act of 1850.
\newblock \emph{The Journal of Law and Economics}, 59(3):669--695.
\newblock Publisher: University of Chicago Press Chicago, IL.

\bibitem[{Levenshtein(1966)}]{levenshtein_binary_1966}
Vladimir~I. Levenshtein. 1966.
\newblock Binary codes capable of correcting deletions, insertions, and reversals.
\newblock In \emph{Soviet physics doklady}, volume~10, pages 707--710. Soviet Union.
\newblock Issue: 8.

\bibitem[{Lloyd(1939)}]{lloyd_slavery_1939}
Arthur~Young Lloyd. 1939.
\newblock \emph{The {Slavery} {Controversy}, 1831-1860}.
\newblock University of North Carolina Press.

\bibitem[{Lorang and Zillig(2012)}]{lorang_electronic_2012}
Elizabeth Lorang and Brian~Pytlik Zillig. 2012.
\newblock Electronic text analysis and nineteenth-century newspapers: {TokenX} and the {Richmond} {Daily} {Dispatch}.
\newblock \emph{Texas Studies in Literature and Language}, 54(3):303--323.
\newblock Publisher: University of Texas Press.

\bibitem[{Malcolm(1990)}]{malcolm1990malcolm}
X~Malcolm. 1990.
\newblock \emph{Malcolm X speaks: Selected speeches and statements}.
\newblock Grove Press.

\bibitem[{McPherson(2003)}]{mcpherson_reconstructing_2003}
Tara McPherson. 2003.
\newblock \emph{Reconstructing {Dixie}: {Race}, gender, and nostalgia in the imagined {South}}.
\newblock Duke University Press.

\bibitem[{Meyer(2017)}]{meyer_economics_2017}
John~R. Meyer. 2017.
\newblock \emph{The {Economics} of {Slavery}: {And} {Other} {Studies} in {Econometric} {History}}.
\newblock Routledge.

\bibitem[{Mikolov et~al.(2013)Mikolov, Sutskever, Chen, Corrado, and Dean}]{mikolov_distributed_2013}
Tomas Mikolov, Ilya Sutskever, Kai Chen, Greg~S. Corrado, and Jeff Dean. 2013.
\newblock Distributed representations of words and phrases and their compositionality.
\newblock \emph{Advances in neural information processing systems}, 26:1--9.

\bibitem[{Monroe et~al.(2008)Monroe, Colaresi, and Quinn}]{monroe2008fightin}
Burt~L Monroe, Michael~P Colaresi, and Kevin~M Quinn. 2008.
\newblock Fightin'words: Lexical feature selection and evaluation for identifying the content of political conflict.
\newblock \emph{Political Analysis}, 16(4):372--403.

\bibitem[{Narayan(2020)}]{narayan_slavery_2020}
Rosalyn Narayan. 2020.
\newblock \emph{Slavery in print: slaveholding ideology and anxiety in antebellum southern newspapers, 1830-1861}.
\newblock Ph.D. thesis, University of Warwick.

\bibitem[{Park and Cordell(2023)}]{park2023quantitative}
Jaihyun Park and Ryan Cordell. 2023.
\newblock A quantitative discourse analysis of asian workers in the us historical newspapers.
\newblock In \emph{The Joint 3rd International Conference on Natural Language Processing for Digital Humanities and 8th International Workshop on Computational Linguistics for Uralic Languages}, page~7.

\bibitem[{Park and Jeoung(2022)}]{park2022raison}
Jaihyun Park and Sullam Jeoung. 2022.
\newblock Raison d’{\^e}tre of the benchmark dataset: A survey of current practices of benchmark dataset sharing platforms.
\newblock In \emph{Proceedings of NLP Power! The First Workshop on Efficient Benchmarking in NLP}, pages 1--10.

\bibitem[{Park et~al.(2024)Park, Yang, Tolbert, and Bunsold}]{park2024you}
Jaihyun Park, JungHwan Yang, Amanda Tolbert, and Katherine Bunsold. 2024.
\newblock You change the way you talk: Examining the network, toxicity and discourse of cross-platform users on twitter and parler during the 2020 us presidential election.
\newblock \emph{Journal of Information Science}, page 01655515241238405.

\bibitem[{Pasley(2002)}]{pasley_tyranny_2002}
Jeffrey~L. Pasley. 2002.
\newblock \emph{The tyranny of printers: {Newspaper} politics in the early {American} republic}.
\newblock University of Virginia Press.

\bibitem[{Rae(2018)}]{rae_great_2018}
Noel Rae. 2018.
\newblock \emph{The {Great} {Stain}: {Witnessing} {American} {Slavery}}.
\newblock Abrams.

\bibitem[{Smith et~al.(2014)Smith, Cordell, Dillon, Stramp, and Wilkerson}]{smith_detecting_2014}
David~A. Smith, Ryan Cordell, Elizabeth~Maddock Dillon, Nick Stramp, and John Wilkerson. 2014.
\newblock \href {https://doi.org/10.1109/JCDL.2014.6970166} {Detecting and modeling local text reuse}.
\newblock In \emph{{IEEE}/{ACM} {Joint} {Conference} on {Digital} {Libraries}}, pages 183--192.

\bibitem[{Smith et~al.(2015)Smith, Cordell, and Mullen}]{smith_computational_2015}
David~A. Smith, Ryan Cordell, and Abby Mullen. 2015.
\newblock \href {https://doi.org/10.1093/alh/ajv029} {Computational {Methods} for {Uncovering} {Reprinted} {Texts} in {Antebellum} {Newspapers}}.
\newblock \emph{American Literary History}, 27(3):E1--E15.

\bibitem[{Soni et~al.(2021)Soni, Klein, and Eisenstein}]{soni2021abolitionist}
Sandeep Soni, Lauren~F Klein, and Jacob Eisenstein. 2021.
\newblock Abolitionist networks: Modeling language change in nineteenth-century activist newspapers.
\newblock \emph{Journal of Cultural Analytics}, 6(1).

\bibitem[{Traub et~al.(2015)Traub, van Ossenbruggen, and Hardman}]{traub_impact_2015}
Myriam~C. Traub, Jacco van Ossenbruggen, and Lynda Hardman. 2015.
\newblock \href {https://doi.org/10.1007/978-3-319-24592-8_19} {Impact {Analysis} of {OCR} {Quality} on {Research} {Tasks} in {Digital} {Archives}}.
\newblock In \emph{Research and {Advanced} {Technology} for {Digital} {Libraries}}, Lecture {Notes} in {Computer} {Science}, pages 252--263, Cham. Springer International Publishing.

\bibitem[{Viera and Garrett(2005)}]{viera_understanding_nodate}
Anthony~J Viera and Joanne~M Garrett. 2005.
\newblock Understanding {Interobserver} {Agreement}: {The} {Kappa} {Statistic}.
\newblock \emph{Family Medicine}.

\bibitem[{Willaert et~al.(2022)Willaert, Banisch, Van~Eecke, and Beuls}]{willaert2022tracking}
Tom Willaert, Sven Banisch, Paul Van~Eecke, and Katrien Beuls. 2022.
\newblock Tracking causal relations in the news: data, tools, and models for the analysis of argumentative statements in online media.

\end{thebibliography}

\appendix

\section{Appendix} \label{sec:appendix A}

\begin{table*}[]
\Large
\resizebox{\textwidth}{!}{%
\begin{tabular}{|c|c|ccccc|}
\hline
\multirow{2}{*}{Stance} & \multirow{2}{*}{Title} & \multicolumn{1}{c|}{\multirow{2}{*}{LCCN}} & \multicolumn{1}{c|}{\multirow{2}{*}{Frequency}} & \multicolumn{1}{c|}{\multirow{2}{*}{Geographic location}} & \multicolumn{2}{c|}{Snippet size} \\ \cline{6-7} 
 &  & \multicolumn{1}{c|}{} & \multicolumn{1}{c|}{} & \multicolumn{1}{c|}{} & \multicolumn{1}{c|}{Servant} & Slave \\ \hline
\multirow{8}{*}{Pro-slavery newspaper} & \multirow{2}{*}{\begin{tabular}[c]{@{}c@{}}Abbeville Banner\\ (1850-1860)\end{tabular}} & \multicolumn{1}{c|}{sn85026945} & \multicolumn{1}{c|}{Weekly} & \multicolumn{1}{c|}{Abbeville, SC} & \multicolumn{1}{c|}{n=314} & n=2,488 \\ \cline{3-7} 
 &  & \multicolumn{5}{c|}{\begin{tabular}[c]{@{}c@{}}Like other white Democrat newspapers of the era,\\  the Press and Banner steered a conservative course, celebrating \\ the return of the ``Bourbons,'' or antebellum-era aristocrats, \\ to political power in 1877 and championing the interests of agrarian elites.\end{tabular}} \\ \cline{2-7} 
 & \multirow{2}{*}{\begin{tabular}[c]{@{}c@{}}Daily Dispatch\\ (1852-1865)\end{tabular}} & \multicolumn{1}{c|}{sn84024738} & \multicolumn{1}{c|}{\begin{tabular}[c]{@{}c@{}}Daily\\ (Except Sundays)\end{tabular}} & \multicolumn{1}{c|}{Richmond, VA} & \multicolumn{1}{c|}{n=17,390} & n=24,975 \\ \cline{3-7} 
 &  & \multicolumn{5}{c|}{\begin{tabular}[c]{@{}c@{}}Though the Daily Dispatch started as nonpartisan, Cowardin, a staunch southern Whig, \\ increasingly included conservative and pro-slavery editorials \\ while advocating the development of local industry \\ as a path to independence at a time of growing sectional tension.\end{tabular}} \\ \cline{2-7} 
 & \multirow{2}{*}{\begin{tabular}[c]{@{}c@{}}Edgefield Advertiser\\ (1850-1862)\end{tabular}} & \multicolumn{1}{c|}{sn84026897} & \multicolumn{1}{c|}{Weekly} & \multicolumn{1}{c|}{Edgefield, SC} & \multicolumn{1}{c|}{n=1,125} & n=11,939 \\ \cline{3-7} 
 &  & \multicolumn{5}{c|}{\begin{tabular}[c]{@{}c@{}}Its editors have at times vigorously defended some of the most divisive issues \\ in this nation's history -- nullification, secession, segregation, slavery, and states' rights.\end{tabular}} \\ \cline{2-7} 
 & \multirow{2}{*}{\begin{tabular}[c]{@{}c@{}}Nashville Union and American\\ (1853-1862)\end{tabular}} & \multicolumn{1}{c|}{sn85038518} & \multicolumn{1}{c|}{\begin{tabular}[c]{@{}c@{}}Daily\\ (Except Mondays)\end{tabular}} & \multicolumn{1}{c|}{Nashville, Davidson, TN} & \multicolumn{1}{c|}{n=6,385} & n=21,721 \\ \cline{3-7} 
 &  & \multicolumn{5}{c|}{\begin{tabular}[c]{@{}c@{}}In the merger announcement on May 17, 1853, \\ the Nashville American assured readers that \\ ``it will be the constant aim of the consolidated journal \\ to preserve the democratic party of Tennessee\\  a unit for all the great purposes of its organization.''\end{tabular}} \\ \hline
\multirow{6}{*}{Anti-slavery newspaper} & \multirow{2}{*}{\begin{tabular}[c]{@{}c@{}}Anti-slavery Bugle\\ (1850-1861)\end{tabular}} & \multicolumn{1}{c|}{sn83035487} & \multicolumn{1}{c|}{Weekly} & \multicolumn{1}{c|}{New-Lisbon, OH} & \multicolumn{1}{c|}{n=1,251} & n=69,877 \\ \cline{3-7} 
 &  & \multicolumn{5}{c|}{\begin{tabular}[c]{@{}c@{}}Marius R. Robin served as editor of the paper for over seven years\\ during the 1850s and was extremely active \\ in the Anti-Slavery Society, once serving as its president.\end{tabular}} \\ \cline{2-7} 
 & \multirow{2}{*}{\begin{tabular}[c]{@{}c@{}}Green Mountain Freeman\\ (1850-1865)\end{tabular}} & \multicolumn{1}{c|}{sn84023209} & \multicolumn{1}{c|}{Weekly} & \multicolumn{1}{c|}{Montpelier, Washington, VT} & \multicolumn{1}{c|}{n=1,097} & n=13,046 \\ \cline{3-7} 
 &  & \multicolumn{5}{c|}{\begin{tabular}[c]{@{}c@{}}From November 1842 to 1843, the Vermont Freeman, published first by antislavery agent \\ and lecturer Alanson St. Clair and then by Joseph E. Hood, with editorial assistance \\ from Chester C. Briggs, was issued from Montpelier and Norwich.\end{tabular}} \\ \cline{2-7} 
 & \multirow{2}{*}{\begin{tabular}[c]{@{}c@{}}National Era\\ (1850-1860)\end{tabular}} & \multicolumn{1}{c|}{sn84026752} & \multicolumn{1}{c|}{Weekly} & \multicolumn{1}{c|}{Washington, DC} & \multicolumn{1}{c|}{n=1,987} & n=37,918 \\ \cline{3-7} 
 &  & \multicolumn{5}{c|}{\begin{tabular}[c]{@{}c@{}}The National Era was an important publisher of abolitionist exists, \\ most notably the serialization of Uncle Tom's Cabin in 1851. \\ Its editor, John Greeleaf Whittier, was a Quaker abolitionist and poet \\ who staunchly advocated for emancipation throughout his time with the Paper.\end{tabular}} \\ \hline
\end{tabular}%
}
\caption{Dataset description.}
\label{tab:data}
\end{table*}

\begin{table*}[]
  \resizebox{\textwidth}{!}{%
  \begin{tabular}{|c|cccc|ccc|}
  \hline
   & \multicolumn{1}{c|}{\begin{tabular}[c]{@{}c@{}}Abbeville\\ Banner\end{tabular}} & \multicolumn{1}{c|}{\begin{tabular}[c]{@{}c@{}}Daily\\ Dispatch\end{tabular}} & \multicolumn{1}{c|}{\begin{tabular}[c]{@{}c@{}}Edgefield\\ Advertiser\end{tabular}} & \begin{tabular}[c]{@{}c@{}}Nashville \\ Union and\\ American\end{tabular} & \multicolumn{1}{c|}{\begin{tabular}[c]{@{}c@{}}Anti-slavery\\ Bugle\end{tabular}} & \multicolumn{1}{c|}{\begin{tabular}[c]{@{}c@{}}Green\\ Mountain\\ Freeman\end{tabular}} & \begin{tabular}[c]{@{}c@{}}National\\ Era\end{tabular} \\ \hline
   & \multicolumn{4}{c|}{Pro-slavery stance newspapers} & \multicolumn{3}{c|}{Anti-slavery stance newspapers} \\ \hline
  1 & \multicolumn{1}{c|}{\begin{tabular}[c]{@{}c@{}}person\\ (0.017)\end{tabular}} & \multicolumn{1}{c|}{\begin{tabular}[c]{@{}c@{}}carry\\ (0.3936)\end{tabular}} & \multicolumn{1}{c|}{\begin{tabular}[c]{@{}c@{}}repeal\\ (0.5490)\end{tabular}} & \begin{tabular}[c]{@{}c@{}}immediate\\ (0.3877)\end{tabular} & \multicolumn{1}{c|}{\begin{tabular}[c]{@{}c@{}}restoration\\ (0.3788)\end{tabular}} & \multicolumn{1}{c|}{\begin{tabular}[c]{@{}c@{}}territory\\ (0.5851)\end{tabular}} & \begin{tabular}[c]{@{}c@{}}gulf\\ (0.4330)\end{tabular} \\ \hline
  2 & \multicolumn{1}{c|}{\begin{tabular}[c]{@{}c@{}}back\\ (0.013)\end{tabular}} & \multicolumn{1}{c|}{\begin{tabular}[c]{@{}c@{}}attempt\\ (0.3670)\end{tabular}} & \multicolumn{1}{c|}{\begin{tabular}[c]{@{}c@{}}territory\\ (0.5415)\end{tabular}} & \begin{tabular}[c]{@{}c@{}}claim\\ (0.3618)\end{tabular} & \multicolumn{1}{c|}{\begin{tabular}[c]{@{}c@{}}entire\\ (0.3591)\end{tabular}} & \multicolumn{1}{c|}{\begin{tabular}[c]{@{}c@{}}admission\\ (0.5846)\end{tabular}} & \begin{tabular}[c]{@{}c@{}}oppose\\ (0.4169)\end{tabular} \\ \hline
  3 & \multicolumn{1}{c|}{\begin{tabular}[c]{@{}c@{}}color\\ (0.010)\end{tabular}} & \multicolumn{1}{c|}{\begin{tabular}[c]{@{}c@{}}death\\ (0.3560)\end{tabular}} & \multicolumn{1}{c|}{\begin{tabular}[c]{@{}c@{}}law\\ (0.5278)\end{tabular}} & \begin{tabular}[c]{@{}c@{}}would\\ (0.3567)\end{tabular} & \multicolumn{1}{c|}{\begin{tabular}[c]{@{}c@{}}acquisition\\ (0.3506)\end{tabular}} & \multicolumn{1}{c|}{\begin{tabular}[c]{@{}c@{}}admit\\ (0.5780)\end{tabular}} & \begin{tabular}[c]{@{}c@{}}favor\\ (0.4017)\end{tabular} \\ \hline
  4 & \multicolumn{1}{c|}{\begin{tabular}[c]{@{}c@{}}population\\ (0.009)\end{tabular}} & \multicolumn{1}{c|}{\begin{tabular}[c]{@{}c@{}}punishment\\ (0.3506)\end{tabular}} & \multicolumn{1}{c|}{\begin{tabular}[c]{@{}c@{}}prohibit\\ (0.5043)\end{tabular}} & \begin{tabular}[c]{@{}c@{}}favor\\ (0.3567)\end{tabular} & \multicolumn{1}{c|}{\begin{tabular}[c]{@{}c@{}}demand\\ (0.3329)\end{tabular}} & \multicolumn{1}{c|}{\begin{tabular}[c]{@{}c@{}}constitution\\ (0.5735)\end{tabular}} & \begin{tabular}[c]{@{}c@{}}interested\\ (0.3940)\end{tabular} \\ \hline
  5 & \multicolumn{1}{c|}{\begin{tabular}[c]{@{}c@{}}profit\\ (0.009)\end{tabular}} & \multicolumn{1}{c|}{\begin{tabular}[c]{@{}c@{}}violation\\ (0.3481)\end{tabular}} & \multicolumn{1}{c|}{\begin{tabular}[c]{@{}c@{}}exclude\\ (0.4961)\end{tabular}} & \begin{tabular}[c]{@{}c@{}}texas\\ (0.3475)\end{tabular} & \multicolumn{1}{c|}{\begin{tabular}[c]{@{}c@{}}native\\ (0.3319)\end{tabular}} & \multicolumn{1}{c|}{\begin{tabular}[c]{@{}c@{}}prohibit\\ (0.5644)\end{tabular}} & \begin{tabular}[c]{@{}c@{}}virginia\\ (0.3940)\end{tabular} \\ \hline
  6 & \multicolumn{1}{c|}{\begin{tabular}[c]{@{}c@{}}say\\ (0.009)\end{tabular}} & \multicolumn{1}{c|}{\begin{tabular}[c]{@{}c@{}}crime\\ (0.3416)\end{tabular}} & \multicolumn{1}{c|}{\begin{tabular}[c]{@{}c@{}}right\\ (0.4944)\end{tabular}} & \begin{tabular}[c]{@{}c@{}}maryland\\ (0.3469)\end{tabular} & \multicolumn{1}{c|}{\begin{tabular}[c]{@{}c@{}}unconditional\\ (0.3262)\end{tabular}} & \multicolumn{1}{c|}{\begin{tabular}[c]{@{}c@{}}missouri\\ (0.5354)\end{tabular}} & \begin{tabular}[c]{@{}c@{}}immigration\\ (0.3928)\end{tabular} \\ \hline
  7 & \multicolumn{1}{c|}{\begin{tabular}[c]{@{}c@{}}upon\\ (0.008)\end{tabular}} & \multicolumn{1}{c|}{\begin{tabular}[c]{@{}c@{}}allege\\ (0.3406)\end{tabular}} & \multicolumn{1}{c|}{\begin{tabular}[c]{@{}c@{}}trade\\ (0.4796)\end{tabular}} & \begin{tabular}[c]{@{}c@{}}proposition\\ (0.3459)\end{tabular} & \multicolumn{1}{c|}{\begin{tabular}[c]{@{}c@{}}enable\\ (0.3252)\end{tabular}} & \multicolumn{1}{c|}{\begin{tabular}[c]{@{}c@{}}exclude\\ (0.5142)\end{tabular}} & \begin{tabular}[c]{@{}c@{}}transfer\\ (0.3902)\end{tabular} \\ \hline
  8 & \multicolumn{1}{c|}{\begin{tabular}[c]{@{}c@{}}land\\ (0.008)\end{tabular}} & \multicolumn{1}{c|}{\begin{tabular}[c]{@{}c@{}}protect\\ (0.3375)\end{tabular}} & \multicolumn{1}{c|}{\begin{tabular}[c]{@{}c@{}}north\\ (0.4767)\end{tabular}} & \begin{tabular}[c]{@{}c@{}}pro\\ (0.3308)\end{tabular} & \multicolumn{1}{c|}{\begin{tabular}[c]{@{}c@{}}stipulation\\ (0.3228)\end{tabular}} & \multicolumn{1}{c|}{\begin{tabular}[c]{@{}c@{}}legislate\\ (0.5141)\end{tabular}} & \begin{tabular}[c]{@{}c@{}}derive\\ (0.3901)\end{tabular} \\ \hline
  9 & \multicolumn{1}{c|}{\begin{tabular}[c]{@{}c@{}}six\\ (0.007)\end{tabular}} & \multicolumn{1}{c|}{\begin{tabular}[c]{@{}c@{}}admit\\ (0.3284)\end{tabular}} & \multicolumn{1}{c|}{\begin{tabular}[c]{@{}c@{}}fugitive\\ (0.4734)\end{tabular}} & \begin{tabular}[c]{@{}c@{}}senator\\ (0.3306)\end{tabular} & \multicolumn{1}{c|}{\begin{tabular}[c]{@{}c@{}}coastwise\\ (0.3217)\end{tabular}} & \multicolumn{1}{c|}{\begin{tabular}[c]{@{}c@{}}united\\ (0.4983)\end{tabular}} & \begin{tabular}[c]{@{}c@{}}similar\\ (0.3845)\end{tabular} \\ \hline
  10 & \multicolumn{1}{c|}{\begin{tabular}[c]{@{}c@{}}labor\\ (0.007)\end{tabular}} & \multicolumn{1}{c|}{\begin{tabular}[c]{@{}c@{}}suppose\\ (0.3264)\end{tabular}} & \multicolumn{1}{c|}{\begin{tabular}[c]{@{}c@{}}union\\ (0.4569)\end{tabular}} & \begin{tabular}[c]{@{}c@{}}justice\\ (0.3303)\end{tabular} & \multicolumn{1}{c|}{\begin{tabular}[c]{@{}c@{}}throughout\\ (0.3207)\end{tabular}} & \multicolumn{1}{c|}{\begin{tabular}[c]{@{}c@{}}establish\\ (0.4972)\end{tabular}} & \begin{tabular}[c]{@{}c@{}}emigration\\ (0.3792)\end{tabular} \\ \hline
  11 & \multicolumn{1}{c|}{\begin{tabular}[c]{@{}c@{}}nation\\ (0.007)\end{tabular}} & \multicolumn{1}{c|}{\begin{tabular}[c]{@{}c@{}}district\\ (0.3218)\end{tabular}} & \multicolumn{1}{c|}{\begin{tabular}[c]{@{}c@{}}abolition\\ (0.4432)\end{tabular}} & \begin{tabular}[c]{@{}c@{}}louisiana\\ (0.3296)\end{tabular} & \multicolumn{1}{c|}{\begin{tabular}[c]{@{}c@{}}annex\\ (0.3190)\end{tabular}} & \multicolumn{1}{c|}{\begin{tabular}[c]{@{}c@{}}limit\\ (0.4958)\end{tabular}} & \begin{tabular}[c]{@{}c@{}}impossible\\ (0.3702)\end{tabular} \\ \hline
  12 & \multicolumn{1}{c|}{\begin{tabular}[c]{@{}c@{}}well\\ (0.007)\end{tabular}} & \multicolumn{1}{c|}{\begin{tabular}[c]{@{}c@{}}commonwealth\\ (0.3167)\end{tabular}} & \multicolumn{1}{c|}{\begin{tabular}[c]{@{}c@{}}legislate\\ (0.4321)\end{tabular}} & \begin{tabular}[c]{@{}c@{}}piracy\\ (0.3251)\end{tabular} & \multicolumn{1}{c|}{\begin{tabular}[c]{@{}c@{}}internal\\ (0.3187)\end{tabular}} & \multicolumn{1}{c|}{\begin{tabular}[c]{@{}c@{}}state\\ (0.4944)\end{tabular}} & \begin{tabular}[c]{@{}c@{}}annex\\ (0.3675)\end{tabular} \\ \hline
  13 & \multicolumn{1}{c|}{\begin{tabular}[c]{@{}c@{}}million\\ (0.006)\end{tabular}} & \multicolumn{1}{c|}{\begin{tabular}[c]{@{}c@{}}matter\\ (0.3158)\end{tabular}} & \multicolumn{1}{c|}{\begin{tabular}[c]{@{}c@{}}constitution\\ (0.4280)\end{tabular}} & \begin{tabular}[c]{@{}c@{}}beyond\\ (0.3233)\end{tabular} & \multicolumn{1}{c|}{\begin{tabular}[c]{@{}c@{}}obtain\\ (0.3173)\end{tabular}} & \multicolumn{1}{c|}{\begin{tabular}[c]{@{}c@{}}district\\ (0.4903)\end{tabular}} & \begin{tabular}[c]{@{}c@{}}prevent\\ (0.3654)\end{tabular} \\ \hline
  14 & \multicolumn{1}{c|}{\begin{tabular}[c]{@{}c@{}}year\\ (0.006)\end{tabular}} & \multicolumn{1}{c|}{\begin{tabular}[c]{@{}c@{}}decision\\ (0.3145)\end{tabular}} & \multicolumn{1}{c|}{\begin{tabular}[c]{@{}c@{}}revive\\ (0.4259)\end{tabular}} & \begin{tabular}[c]{@{}c@{}}attempt\\ (0.3203)\end{tabular} & \multicolumn{1}{c|}{\begin{tabular}[c]{@{}c@{}}restore\\ (0.3142)\end{tabular}} & \multicolumn{1}{c|}{\begin{tabular}[c]{@{}c@{}}existence\\ (0.4876)\end{tabular}} & \begin{tabular}[c]{@{}c@{}}tennessee\\ (0.3585)\end{tabular} \\ \hline
  15 & \multicolumn{1}{c|}{\begin{tabular}[c]{@{}c@{}}cause\\ (0.006)\end{tabular}} & \multicolumn{1}{c|}{\begin{tabular}[c]{@{}c@{}}marshal\\ (0.3143)\end{tabular}} & \multicolumn{1}{c|}{\begin{tabular}[c]{@{}c@{}}exist\\ (0.4254)\end{tabular}} & \begin{tabular}[c]{@{}c@{}}effect\\ (0.3197)\end{tabular} & \multicolumn{1}{c|}{\begin{tabular}[c]{@{}c@{}}uni\\ (0.3136)\end{tabular}} & \multicolumn{1}{c|}{\begin{tabular}[c]{@{}c@{}}exist\\ (0.4860)\end{tabular}} & \begin{tabular}[c]{@{}c@{}}gain\\ (0.3549)\end{tabular} \\ \hline
  16 & \multicolumn{1}{c|}{\begin{tabular}[c]{@{}c@{}}owner\\ (0.005)\end{tabular}} & \multicolumn{1}{c|}{\begin{tabular}[c]{@{}c@{}}concern\\ (0.3141)\end{tabular}} & \multicolumn{1}{c|}{\begin{tabular}[c]{@{}c@{}}reopen\\ (0.4092)\end{tabular}} & \begin{tabular}[c]{@{}c@{}}demand\\ (0.3195)\end{tabular} & \multicolumn{1}{c|}{\begin{tabular}[c]{@{}c@{}}suppression\\ (0.3105)\end{tabular}} & \multicolumn{1}{c|}{\begin{tabular}[c]{@{}c@{}}congress\\ (0.4761)\end{tabular}} & \begin{tabular}[c]{@{}c@{}}include\\ (0.3541)\end{tabular} \\ \hline
  17 & \multicolumn{1}{c|}{\begin{tabular}[c]{@{}c@{}}african\\ (0.005)\end{tabular}} & \multicolumn{1}{c|}{\begin{tabular}[c]{@{}c@{}}pro\\ (0.3140)\end{tabular}} & \multicolumn{1}{c|}{\begin{tabular}[c]{@{}c@{}}african\\ (0.4083)\end{tabular}} & \begin{tabular}[c]{@{}c@{}}especially\\ (0.3190)\end{tabular} & \multicolumn{1}{c|}{\begin{tabular}[c]{@{}c@{}}acquiesce\\ (0.3105)\end{tabular}} & \multicolumn{1}{c|}{\begin{tabular}[c]{@{}c@{}}free\\ (0.4755)\end{tabular}} & \begin{tabular}[c]{@{}c@{}}total\\ (0.3539)\end{tabular} \\ \hline
  18 & \multicolumn{1}{c|}{\begin{tabular}[c]{@{}c@{}}foreign\\ (0.005)\end{tabular}} & \multicolumn{1}{c|}{\begin{tabular}[c]{@{}c@{}}prevent\\ (0.3131)\end{tabular}} & \multicolumn{1}{c|}{\begin{tabular}[c]{@{}c@{}}congress\\ (0.4041)\end{tabular}} & \begin{tabular}[c]{@{}c@{}}labor\\ (0.3188)\end{tabular} & \multicolumn{1}{c|}{\begin{tabular}[c]{@{}c@{}}piratical\\ (0.3064)\end{tabular}} & \multicolumn{1}{c|}{\begin{tabular}[c]{@{}c@{}}prohibition\\ (0.4761)\end{tabular}} & \begin{tabular}[c]{@{}c@{}}demand\\ (0.3538)\end{tabular} \\ \hline
  19 & \multicolumn{1}{c|}{\begin{tabular}[c]{@{}c@{}}answer\\ (0.004)\end{tabular}} & \multicolumn{1}{c|}{\begin{tabular}[c]{@{}c@{}}account\\ (0.3123)\end{tabular}} & \multicolumn{1}{c|}{\begin{tabular}[c]{@{}c@{}}foreign\\ (0.4036)\end{tabular}} & \begin{tabular}[c]{@{}c@{}}intend\\ (0.3176)\end{tabular} & \multicolumn{1}{c|}{\begin{tabular}[c]{@{}c@{}}supremacy\\ (0.3058)\end{tabular}} & \multicolumn{1}{c|}{\begin{tabular}[c]{@{}c@{}}within\\ (0.4691)\end{tabular}} & \begin{tabular}[c]{@{}c@{}}interdict\\ (0.3505)\end{tabular} \\ \hline
  20 & \multicolumn{1}{c|}{\begin{tabular}[c]{@{}c@{}}also\\ (0.004)\end{tabular}} & \multicolumn{1}{c|}{\begin{tabular}[c]{@{}c@{}}resist\\ (0.3114)\end{tabular}} & \multicolumn{1}{c|}{\begin{tabular}[c]{@{}c@{}}carry\\ (0.3984)\end{tabular}} & \begin{tabular}[c]{@{}c@{}}states\\ (0.3175)\end{tabular} & \multicolumn{1}{c|}{\begin{tabular}[c]{@{}c@{}}defiance\\ (0.3049)\end{tabular}} & \multicolumn{1}{c|}{\begin{tabular}[c]{@{}c@{}}introduction\\ (0.4657)\end{tabular}} & \begin{tabular}[c]{@{}c@{}}encourage\\ (0.3489)\end{tabular} \\ \hline
  \end{tabular}%
  }
  \caption{The cosine similarity ranking after deducting ``servant'' from ``slave''}
  \label{tab:my-result-slave}
  \end{table*}

  \begin{table*}[]
    \resizebox{\textwidth}{!}{%
    \begin{tabular}{|c|cccc|ccc|}
    \hline
     & \multicolumn{1}{c|}{\begin{tabular}[c]{@{}c@{}}Abbeville\\ Banner\end{tabular}} & \multicolumn{1}{c|}{\begin{tabular}[c]{@{}c@{}}Daily\\ Dispatch\end{tabular}} & \multicolumn{1}{c|}{\begin{tabular}[c]{@{}c@{}}Edgefield\\ Advertiser\end{tabular}} & \begin{tabular}[c]{@{}c@{}}Nashville \\ Union and\\ American\end{tabular} & \multicolumn{1}{c|}{\begin{tabular}[c]{@{}c@{}}Anti-slavery\\ Bugle\end{tabular}} & \multicolumn{1}{c|}{\begin{tabular}[c]{@{}c@{}}Green\\ Mountain\\ Freeman\end{tabular}} & \begin{tabular}[c]{@{}c@{}}National\\ Era\end{tabular} \\ \hline
     & \multicolumn{4}{c|}{Pro-slavery stance newspapers} & \multicolumn{3}{c|}{Anti-slavery stance newspapers} \\ \hline
    1 & \multicolumn{1}{c|}{\begin{tabular}[c]{@{}c@{}}vote\\ (0.023)\end{tabular}} & \multicolumn{1}{c|}{\begin{tabular}[c]{@{}c@{}}table\\ (0.3568)\end{tabular}} & \multicolumn{1}{c|}{\begin{tabular}[c]{@{}c@{}}respectfully\\ (0.4777)\end{tabular}} & \begin{tabular}[c]{@{}c@{}}linen\\ (0.2942)\end{tabular} & \multicolumn{1}{c|}{\begin{tabular}[c]{@{}c@{}}thy\\ (0.1893)\end{tabular}} & \multicolumn{1}{c|}{\begin{tabular}[c]{@{}c@{}}child\\ (0.3562)\end{tabular}} & \begin{tabular}[c]{@{}c@{}}obedient\\ (0.3312)\end{tabular} \\ \hline
    2 & \multicolumn{1}{c|}{\begin{tabular}[c]{@{}c@{}}party\\ (0.022)\end{tabular}} & \multicolumn{1}{c|}{\begin{tabular}[c]{@{}c@{}}moat\\ (0.3532)\end{tabular}} & \multicolumn{1}{c|}{\begin{tabular}[c]{@{}c@{}}obedient\\ (0.4628)\end{tabular}} & \begin{tabular}[c]{@{}c@{}}good\\ (0.2801)\end{tabular} & \multicolumn{1}{c|}{\begin{tabular}[c]{@{}c@{}}lecture\\ (0.1885)\end{tabular}} & \multicolumn{1}{c|}{\begin{tabular}[c]{@{}c@{}}see\\ (0.3460)\end{tabular}} & \begin{tabular}[c]{@{}c@{}}thy\\ (0.2840)\end{tabular} \\ \hline
    3 & \multicolumn{1}{c|}{\begin{tabular}[c]{@{}c@{}}proslavery\\ (0.021)\end{tabular}} & \multicolumn{1}{c|}{\begin{tabular}[c]{@{}c@{}}gentleman\\ (0.3528)\end{tabular}} & \multicolumn{1}{c|}{\begin{tabular}[c]{@{}c@{}}friend\\ (0.4609)\end{tabular}} & \begin{tabular}[c]{@{}c@{}}bed\\ (0.2740)\end{tabular} & \multicolumn{1}{c|}{\begin{tabular}[c]{@{}c@{}}wife\\ (0.1741)\end{tabular}} & \multicolumn{1}{c|}{\begin{tabular}[c]{@{}c@{}}old\\ (0.3344)\end{tabular}} & \begin{tabular}[c]{@{}c@{}}respectfully\\ (0.2479)\end{tabular} \\ \hline
    4 & \multicolumn{1}{c|}{\begin{tabular}[c]{@{}c@{}}whole\\ (0.021)\end{tabular}} & \multicolumn{1}{c|}{\begin{tabular}[c]{@{}c@{}}ice\\ (0.3412)\end{tabular}} & \multicolumn{1}{c|}{\begin{tabular}[c]{@{}c@{}}seat\\ (0.4422)\end{tabular}} & \begin{tabular}[c]{@{}c@{}}sound\\ (0.2583)\end{tabular} & \multicolumn{1}{c|}{\begin{tabular}[c]{@{}c@{}}master\\ (0.1712)\end{tabular}} & \multicolumn{1}{c|}{\begin{tabular}[c]{@{}c@{}}thy\\ (0.3089)\end{tabular}} & \begin{tabular}[c]{@{}c@{}}god\\ (0.2214)\end{tabular} \\ \hline
    5 & \multicolumn{1}{c|}{\begin{tabular}[c]{@{}c@{}}liberal\\ (0.0207)\end{tabular}} & \multicolumn{1}{c|}{\begin{tabular}[c]{@{}c@{}}furniture\\ (0.3384)\end{tabular}} & \multicolumn{1}{c|}{\begin{tabular}[c]{@{}c@{}}john\\ (0.4367)\end{tabular}} & \begin{tabular}[c]{@{}c@{}}shirt\\ (0.2583)\end{tabular} & \multicolumn{1}{c|}{\begin{tabular}[c]{@{}c@{}}jesus\\ (0.1674)\end{tabular}} & \multicolumn{1}{c|}{\begin{tabular}[c]{@{}c@{}}girl\\ (0.2928)\end{tabular}} & \begin{tabular}[c]{@{}c@{}}brother\\ (0.2178)\end{tabular} \\ \hline
    6 & \multicolumn{1}{c|}{\begin{tabular}[c]{@{}c@{}}allegiance\\ (0.018)\end{tabular}} & \multicolumn{1}{c|}{\begin{tabular}[c]{@{}c@{}}dry\\ (0.3369)\end{tabular}} & \multicolumn{1}{c|}{\begin{tabular}[c]{@{}c@{}}announce\\ (0.4275)\end{tabular}} & \begin{tabular}[c]{@{}c@{}}dress\\ (0.2560)\end{tabular} & \multicolumn{1}{c|}{\begin{tabular}[c]{@{}c@{}}unto\\ (0.1667)\end{tabular}} & \multicolumn{1}{c|}{\begin{tabular}[c]{@{}c@{}}obedient\\ (0.2884)\end{tabular}} & \begin{tabular}[c]{@{}c@{}}honor\\ (0.2071)\end{tabular} \\ \hline
    7 & \multicolumn{1}{c|}{\begin{tabular}[c]{@{}c@{}}commit\\ (0.017)\end{tabular}} & \multicolumn{1}{c|}{\begin{tabular}[c]{@{}c@{}}superior\\ (0.3216)\end{tabular}} & \multicolumn{1}{c|}{\begin{tabular}[c]{@{}c@{}}candidate\\ (0.4056)\end{tabular}} & \begin{tabular}[c]{@{}c@{}}cook\\ (0.2517)\end{tabular} & \multicolumn{1}{c|}{\begin{tabular}[c]{@{}c@{}}song\\ (0.1578)\end{tabular}} & \multicolumn{1}{c|}{\begin{tabular}[c]{@{}c@{}}brother\\ (0.2802)\end{tabular}} & \begin{tabular}[c]{@{}c@{}}christ\\ (0.2062)\end{tabular} \\ \hline
    8 & \multicolumn{1}{c|}{\begin{tabular}[c]{@{}c@{}}slavery\\ (0.016)\end{tabular}} & \multicolumn{1}{c|}{\begin{tabular}[c]{@{}c@{}}good\\ (0.3204)\end{tabular}} & \multicolumn{1}{c|}{\begin{tabular}[c]{@{}c@{}}george\\ (0.4011)\end{tabular}} & \begin{tabular}[c]{@{}c@{}}excellent\\ (0.2514)\end{tabular} & \multicolumn{1}{c|}{\begin{tabular}[c]{@{}c@{}}mind\\ (0.1541)\end{tabular}} & \multicolumn{1}{c|}{\begin{tabular}[c]{@{}c@{}}day\\ (0.2739)\end{tabular}} & \begin{tabular}[c]{@{}c@{}}humble\\ (0.1831)\end{tabular} \\ \hline
    9 & \multicolumn{1}{c|}{\begin{tabular}[c]{@{}c@{}}servile\\ (0.016)\end{tabular}} & \multicolumn{1}{c|}{\begin{tabular}[c]{@{}c@{}}price\\ (0.3196)\end{tabular}} & \multicolumn{1}{c|}{\begin{tabular}[c]{@{}c@{}}ensue\\ (0.3895)\end{tabular}} & \begin{tabular}[c]{@{}c@{}}train\\ (0.2507)\end{tabular} & \multicolumn{1}{c|}{\begin{tabular}[c]{@{}c@{}}doctor\\ (0.1406)\end{tabular}} & \multicolumn{1}{c|}{\begin{tabular}[c]{@{}c@{}}mother\\ (0.2654)\end{tabular}} & \begin{tabular}[c]{@{}c@{}}bible\\ (0.1827)\end{tabular} \\ \hline
    10 & \multicolumn{1}{c|}{\begin{tabular}[c]{@{}c@{}}last\\ (0.016)\end{tabular}} & \multicolumn{1}{c|}{\begin{tabular}[c]{@{}c@{}}rood\\ (0.3154)\end{tabular}} & \multicolumn{1}{c|}{\begin{tabular}[c]{@{}c@{}}cook\\ (0.3749)\end{tabular}} & \begin{tabular}[c]{@{}c@{}}mill\\ (0.2467)\end{tabular} & \multicolumn{1}{c|}{\begin{tabular}[c]{@{}c@{}}mother\\ (0.1385)\end{tabular}} & \multicolumn{1}{c|}{\begin{tabular}[c]{@{}c@{}}respectfully\\ (0.2651)\end{tabular}} & \begin{tabular}[c]{@{}c@{}}thou\\ (0.1773)\end{tabular} \\ \hline
    11 & \multicolumn{1}{c|}{\begin{tabular}[c]{@{}c@{}}arm\\ (0.016)\end{tabular}} & \multicolumn{1}{c|}{\begin{tabular}[c]{@{}c@{}}season\\ (0.3143)\end{tabular}} & \multicolumn{1}{c|}{\begin{tabular}[c]{@{}c@{}}house\\ (0.3738)\end{tabular}} & \begin{tabular}[c]{@{}c@{}}ticking\\ (0.2464)\end{tabular} & \multicolumn{1}{c|}{\begin{tabular}[c]{@{}c@{}}child\\ (0.1360)\end{tabular}} & \multicolumn{1}{c|}{\begin{tabular}[c]{@{}c@{}}live\\ (0.2647)\end{tabular}} & \begin{tabular}[c]{@{}c@{}}heart\\ (0.1766)\end{tabular} \\ \hline
    12 & \multicolumn{1}{c|}{\begin{tabular}[c]{@{}c@{}}desire\\ (0.016)\end{tabular}} & \multicolumn{1}{c|}{\begin{tabular}[c]{@{}c@{}}hoy\\ (0.3100)\end{tabular}} & \multicolumn{1}{c|}{\begin{tabular}[c]{@{}c@{}}james\\ (0.3732)\end{tabular}} & \begin{tabular}[c]{@{}c@{}}flannel\\ (0.2554)\end{tabular} & \multicolumn{1}{c|}{\begin{tabular}[c]{@{}c@{}}mas\\ (0.1349)\end{tabular}} & \multicolumn{1}{c|}{\begin{tabular}[c]{@{}c@{}}good\\ (0.2641)\end{tabular}} & \begin{tabular}[c]{@{}c@{}}chain\\ (0.1706)\end{tabular} \\ \hline
    13 & \multicolumn{1}{c|}{\begin{tabular}[c]{@{}c@{}}owe\\ (0.016)\end{tabular}} & \multicolumn{1}{c|}{\begin{tabular}[c]{@{}c@{}}garden\\ (0.3076)\end{tabular}} & \multicolumn{1}{c|}{\begin{tabular}[c]{@{}c@{}}jones\\ (0.3614)\end{tabular}} & \begin{tabular}[c]{@{}c@{}}apron\\ (0.2447)\end{tabular} & \multicolumn{1}{c|}{\begin{tabular}[c]{@{}c@{}}christ\\ (0.1348)\end{tabular}} & \multicolumn{1}{c|}{\begin{tabular}[c]{@{}c@{}}woman\\ (0.2576)\end{tabular}} & \begin{tabular}[c]{@{}c@{}}unto\\ (0.1676)\end{tabular} \\ \hline
    14 & \multicolumn{1}{c|}{\begin{tabular}[c]{@{}c@{}}institution\\ (0.015)\end{tabular}} & \multicolumn{1}{c|}{\begin{tabular}[c]{@{}c@{}}summer\\ (0.3075)\end{tabular}} & \multicolumn{1}{c|}{\begin{tabular}[c]{@{}c@{}}reelection\\ (0.3614)\end{tabular}} & \begin{tabular}[c]{@{}c@{}}cottonade\\ (0.2426)\end{tabular} & \multicolumn{1}{c|}{\begin{tabular}[c]{@{}c@{}}husband\\ (0.1315)\end{tabular}} & \multicolumn{1}{c|}{\begin{tabular}[c]{@{}c@{}}god\\ (0.2567)\end{tabular}} & \begin{tabular}[c]{@{}c@{}}faithful\\ (0.1599)\end{tabular} \\ \hline
    15 & \multicolumn{1}{c|}{\begin{tabular}[c]{@{}c@{}}measure\\ (0.015)\end{tabular}} & \multicolumn{1}{c|}{\begin{tabular}[c]{@{}c@{}}band\\ (0.3075)\end{tabular}} & \multicolumn{1}{c|}{\begin{tabular}[c]{@{}c@{}}abraham\\ (0.3549)\end{tabular}} & \begin{tabular}[c]{@{}c@{}}blanket\\ (0.2425)\end{tabular} & \multicolumn{1}{c|}{\begin{tabular}[c]{@{}c@{}}book\\ (0.1291)\end{tabular}} & \multicolumn{1}{c|}{\begin{tabular}[c]{@{}c@{}}two\\ (0.2527)\end{tabular}} & \begin{tabular}[c]{@{}c@{}}church\\ (0.1572)\end{tabular} \\ \hline
    16 & \multicolumn{1}{c|}{\begin{tabular}[c]{@{}c@{}}authority\\ (0.015)\end{tabular}} & \multicolumn{1}{c|}{\begin{tabular}[c]{@{}c@{}}ladle\\ (0.3018)\end{tabular}} & \multicolumn{1}{c|}{\begin{tabular}[c]{@{}c@{}}representative\\ (0.3531)\end{tabular}} & \begin{tabular}[c]{@{}c@{}}solicit\\ (0.2422)\end{tabular} & \multicolumn{1}{c|}{\begin{tabular}[c]{@{}c@{}}cony\\ (0.1283)\end{tabular}} & \multicolumn{1}{c|}{\begin{tabular}[c]{@{}c@{}}know\\ (0.2502)\end{tabular}} & \begin{tabular}[c]{@{}c@{}}poor\\ (0.1566)\end{tabular} \\ \hline
    17 & \multicolumn{1}{c|}{\begin{tabular}[c]{@{}c@{}}prohibit\\ (0.015)\end{tabular}} & \multicolumn{1}{c|}{\begin{tabular}[c]{@{}c@{}}excellent\\ (0.3007)\end{tabular}} & \multicolumn{1}{c|}{\begin{tabular}[c]{@{}c@{}}mar\\ (0.3377)\end{tabular}} & \begin{tabular}[c]{@{}c@{}}unusually\\ (0.2414)\end{tabular} & \multicolumn{1}{c|}{\begin{tabular}[c]{@{}c@{}}away\\ (0.1252)\end{tabular}} & \multicolumn{1}{c|}{\begin{tabular}[c]{@{}c@{}}family\\ (0.2472)\end{tabular}} & \begin{tabular}[c]{@{}c@{}}girl\\ (0.1533)\end{tabular} \\ \hline
    18 & \multicolumn{1}{c|}{\begin{tabular}[c]{@{}c@{}}excitement\\ (0.015)\end{tabular}} & \multicolumn{1}{c|}{\begin{tabular}[c]{@{}c@{}}hood\\ (0.2992)\end{tabular}} & \multicolumn{1}{c|}{\begin{tabular}[c]{@{}c@{}}scat\\ (0.3239)\end{tabular}} & \begin{tabular}[c]{@{}c@{}}hue\\ (0.2398)\end{tabular} & \multicolumn{1}{c|}{\begin{tabular}[c]{@{}c@{}}obedient\\ (0.1186)\end{tabular}} & \multicolumn{1}{c|}{\begin{tabular}[c]{@{}c@{}}poor\\ (0.2446)\end{tabular}} & \begin{tabular}[c]{@{}c@{}}write\\ (0.1526)\end{tabular} \\ \hline
    19 & \multicolumn{1}{c|}{\begin{tabular}[c]{@{}c@{}}judge\\ (0.015)\end{tabular}} & \multicolumn{1}{c|}{\begin{tabular}[c]{@{}c@{}}children\\ (0.2991)\end{tabular}} & \multicolumn{1}{c|}{\begin{tabular}[c]{@{}c@{}}nominate\\ (0.3204)\end{tabular}} & \begin{tabular}[c]{@{}c@{}}coarse\\ (0.2373)\end{tabular} & \multicolumn{1}{c|}{\begin{tabular}[c]{@{}c@{}}bible\\ (0.1184)\end{tabular}} & \multicolumn{1}{c|}{\begin{tabular}[c]{@{}c@{}}wife\\ (0.2398)\end{tabular}} & \begin{tabular}[c]{@{}c@{}}father\\ (0.1442)\end{tabular} \\ \hline
    20 & \multicolumn{1}{c|}{\begin{tabular}[c]{@{}c@{}}like\\ (0.014)\end{tabular}} & \multicolumn{1}{c|}{\begin{tabular}[c]{@{}c@{}}dress\\ (0.2979)\end{tabular}} & \multicolumn{1}{c|}{\begin{tabular}[c]{@{}c@{}}many\\ (0.2951)\end{tabular}} & \begin{tabular}[c]{@{}c@{}}men\\ (0.2358)\end{tabular} & \multicolumn{1}{c|}{\begin{tabular}[c]{@{}c@{}}sin\\ (0.1157)\end{tabular}} & \multicolumn{1}{c|}{\begin{tabular}[c]{@{}c@{}}like\\ (0.2366)\end{tabular}} & \begin{tabular}[c]{@{}c@{}}wife\\ (0.1425)\end{tabular} \\ \hline
    \end{tabular}%
    }
    \caption{The cosine similarity ranking after deducting ``slave'' from ``servant''}
    \label{tab:my-result-servant}
  \end{table*}



\end{document}